\documentclass{article}

\usepackage{PRIMEarxiv}

\usepackage[utf8]{inputenc} 
\usepackage[T1]{fontenc}    
\usepackage{hyperref}       
\usepackage{url}            
\usepackage{booktabs}       
\usepackage{amsfonts}       
\usepackage{nicefrac}       
\usepackage{microtype}      
\usepackage{lipsum}
\usepackage{fancyhdr}       
\usepackage{graphicx}       
\graphicspath{{media/}} 
\usepackage{amsmath}
\title{Kolmogorov-Arnold Networks for Time Series: Bridging Predictive Power and Interpretability

}

\author{
  Kunpeng Xu, Lifei Chen, Shengrui Wang \\
  Université de Sherbrooke \\
  Québec, Canada
}

\begin{document}
\maketitle

\begin{abstract}
Kolmogorov-Arnold Networks (KAN) is a groundbreaking model recently proposed by the MIT team, representing a revolutionary approach with the potential to be a game-changer in the field. This innovative concept has rapidly garnered worldwide interest within the AI community. Inspired by the Kolmogorov-Arnold representation theorem, KAN utilizes spline-parametrized univariate functions in place of traditional linear weights, enabling them to dynamically learn activation patterns and significantly enhancing interpretability. In this paper, we explore the application of KAN to time series forecasting and propose two variants: T-KAN and MT-KAN. T-KAN is designed to detect concept drift within time series and can explain the nonlinear relationships between predictions and previous time steps through symbolic regression, making it highly interpretable in dynamically changing environments. MT-KAN, on the other hand, improves predictive performance by effectively uncovering and leveraging the complex relationships among variables in multivariate time series. Experiments validate the effectiveness of these approaches, demonstrating that T-KAN and MT-KAN significantly outperform traditional methods in time series forecasting tasks, not only enhancing predictive accuracy but also improving model interpretability. This research opens new avenues for adaptive forecasting models, highlighting the potential of KAN as a powerful and interpretable tool in predictive analytics.
\end{abstract}


\section{Introduction}
Time series forecasting is crucial across various fields, such as finance, healthcare, and environmental science. Recently, deep learning models have achieved remarkable success in forecasting tasks. Notable advancements include the Channel Aligned Robust Blend Transformer (CARD) \cite{wang2023card}, which aligns channels within a transformer model to enhance accuracy, and TimeMixer \cite{wang2024timemixer}, which employs a decomposable multiscale mixing approach. MSGNet \cite{cai2024msgnet} leverages multi-scale inter-series correlations for multivariate forecasting, while Crossformer \cite{zhang2022crossformer} exploits cross-dimensional dependencies. Additionally, DeepTime \cite{woo2023learning} integrates time indices into the model structure, improving forecasting outcomes. In parallel, large language models (LLMs) have also been integrated into time series forecasting \cite{jin2023time,liu2024unitime,gruver2024large}, opening new avenues. For instance, Time-LLM \cite{jin2023time} reprograms LLMs for forecasting tasks, demonstrating zero-shot capabilities, while UniTime \cite{liu2024unitime} uses language models for cross-domain forecasting.

\begin{figure}[t]
\centerline{\includegraphics[width=0.8\linewidth]{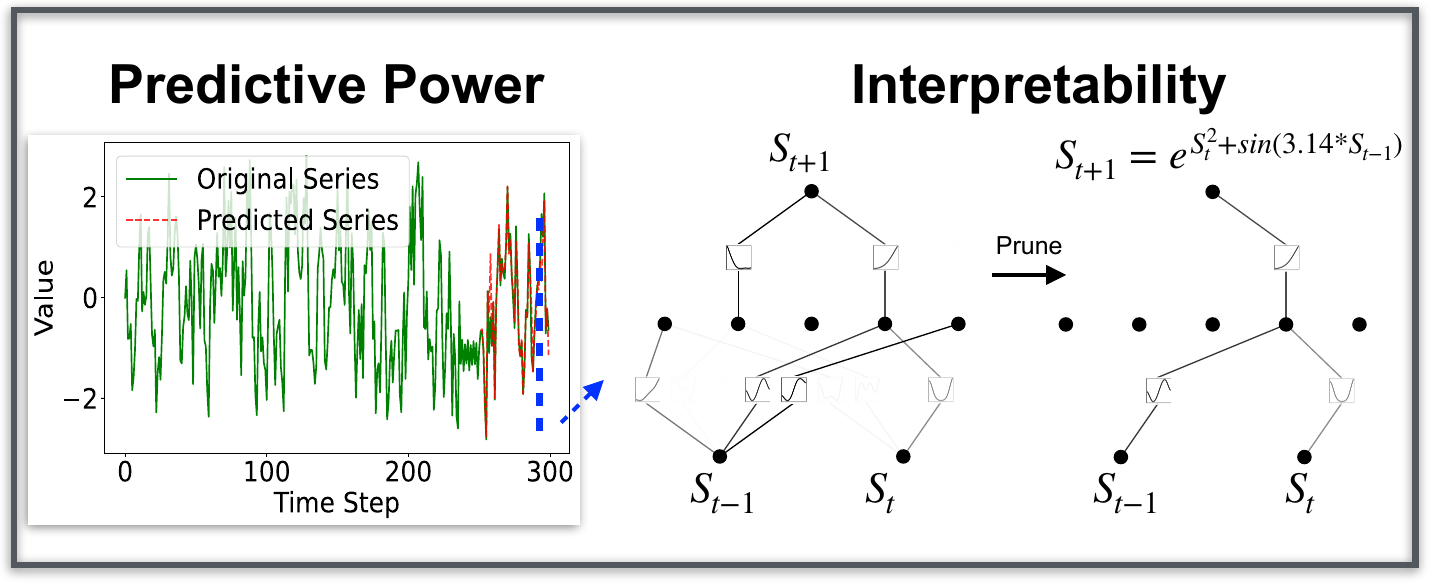}}
\caption{Predictive and Interpretable Capabilities of KAN in Time Series.}
\label{Fig1}
\vspace{-15pt}
\end{figure}
Despite these advancements, interpretability remains a significant challenge for models based on the Multi-Layer Perceptrons (MLPs) architecture \cite{he2020extract,radenovic2022neural}. These models often suffer from poor scaling laws—meaning the number of parameters does not scale linearly with the number of layers—and generally lack interpretability. This lack of transparency is particularly problematic in domains where understanding the underlying patterns and mechanisms is crucial. For example, in financial markets, investors prioritize understanding regime switches and uncovering potential nonlinear relationships among stock series over predictive accuracy \cite{xu2024rhine,xu2024kernel}. Similarly, in other decision-making fields such as meteorology and healthcare, identifying and detecting concept drift in time series is also critical \cite{li2024concept}. Recent efforts, such as OneNet \cite{wen2024onenet} and FSNet \cite{pham2022learning}, have begun to address concept drift in time series forecasting. These methods primarily focus on mitigating the effects of concept drift through online ensembling and learning fast and slow patterns, respectively. However, these approaches do not fully address the challenge of adaptive concept identification and dynamic concept drift in time series.

Kolmogorov-Arnold Networks (KAN) represent a recent innovation proposed by the MIT team \cite{liu2024kan}, published just a month before the submission of this paper. KAN is a novel neural network architecture designed as a potential alternative to MLPs, inspired by the Kolmogorov-Arnold representation theorem \cite{kolmogorov1961representation,kolmogorov1957representation,braun2009constructive}. Unlike MLPs, KAN applies activation functions on the connections between nodes, with these functions being capable of learning and adapting during training. By replacing linear weights with spline-parametrized univariate functions along the network edges, KAN enhances both the accuracy and interpretability of the network. A significant advantage of KAN is that a spline can be made arbitrarily accurate to a target function by refining the grid. This design not only improves network performance but also enables them to achieve comparable or superior results with smaller network sizes across various tasks \cite{liu2024kan,vaca2024kolmogorov,genet2024tkan}. 

In this paper, we conduct a prospective study exploring the application of KAN to time series forecasting and introduce two variants: T-KAN and MT-KAN. T-KAN is designed to monitor concept drift within time series, maintaining high predictive accuracy in dynamically changing environments. It also employs symbolic regression to decode the nonlinear relationships between predictions and prior time steps, significantly enhancing interpretability, see Fig.\ref{Fig1}. Conversely, MT-KAN is designed to effectively identify and leverage complex interdependencies among variables in multivariate time series. Our experimental results on financial time series data reveal that both T-KAN and MT-KAN, with only 2 layers and 5 hidden neurons, achieve comparable results to other MLP-based baselines. 

\section{Kolmogorov-Arnold Networks}
Kolmogorov-Arnold Networks (KAN) is a novel neural network architecture that builds on the Kolmogorov-Arnold representation theorem \cite{kolmogorov1961representation,kolmogorov1957representation,braun2009constructive}. This theorem provides a theoretical foundation for expressing multivariate continuous functions through compositions of univariate functions., allowing for the decomposition of complex functions into simpler, more interpretable parts.

\subsection{Theoretical Foundation}

The Kolmogorov-Arnold representation theorem states that any multivariate continuous function can be decomposed into a finite sum of compositions of univariate functions \cite{kolmogorov1961representation}. Formally, the theorem is expressed as:

\begin{equation}
f(x_1, \ldots, x_n) = \sum_{q=1}^{2n+1} \Phi_q \left( \sum_{p=1}^{n} \varphi_{q,p}(x_p) \right),
\end{equation}

where \( \varphi_{q,p} \) are univariate functions that map each input variable \( x_p \), and \( \Phi_q \) are continuous functions. This allows KAN to model complex interactions in high-dimensional data through compositions of simpler univariate functions.

\subsection{Network Architecture}
KAN leverages the Kolmogorov-Arnold representation theorem by replacing traditional linear weights in neural networks with spline-parametrized univariate functions. Unlike conventional Multi-Layer Perceptrons (MLPs), which use fixed activation functions at the nodes, KAN applies adaptive, learnable activation functions on the edges between nodes. These functions are parametrized as B-spline curves, which adjust dynamically during training to better capture the underlying data patterns. This unique structure enables KAN to effectively capture complex nonlinear relationships within the data. Formally, a KAN layer can be defined as $\Phi = \{ \varphi_{q,p} \}, \quad p = 1, 2, \ldots, n_{\text{in}}, \quad q = 1, 2, \ldots, n_{\text{out}},$, where \( \varphi_{q,p} \) are parametrized functions with learnable parameters. This structure allows KAN to capture complex nonlinear relationships within the data more effectively than traditional Multi-Layer Perceptrons (MLPs).

To extend the capabilities of KAN, deeper network architectures have been developed. A deeper KAN is essentially a composition of multiple KAN layers \cite{liu2024kan}, enhancing its ability to model more complex functions. The architecture of a deeper KAN can be described as:

\begin{equation}
KAN(x) = (\Phi_{L-1} \circ \Phi_{L-2} \circ \ldots \circ \Phi_0)(x),
\end{equation}

where each \( \Phi_l \) represents a KAN layer. The depth of the network (i.e., the number of layers) allows it to capture more intricate patterns and dependencies in the data. Each layer \( l \) transforms the input \( x \) through a series of learnable functions \( \varphi_{q,p} \), making the network highly adaptable and powerful.

\section{KAN for Time Series}
We introduce our proposed models: Temporal Kolmogorov-Arnold Networks (T-KAN) and Multivariate Temporal Kolmogorov-Arnold Networks (MT-KAN) in this section. These models apply the KAN framework to address specific challenges in univariate and multivariate time series forecasting.

\noindent \textbf{Problem Defintion}
Given a time series \(\{S_{t-h}, \ldots, S_t\}\) (univariate) or \(\{\mathbf{S}_{t-h}, \ldots, \mathbf{S}_t\}\) (multivariate), where \(\mathbf{S}_t = (S_{t,1}, \ldots, S_{t,m})\) represents \(m\) variables at time \(t\), our objective is to predict their future values over the next \(T\) time steps. In this work, we transformer the time series forecasting problem as a supervised learning task, where the input-output pairs consist of historical data and the forecasting horizon.

\begin{figure}[t]
\centerline{\includegraphics[width=0.8\linewidth]{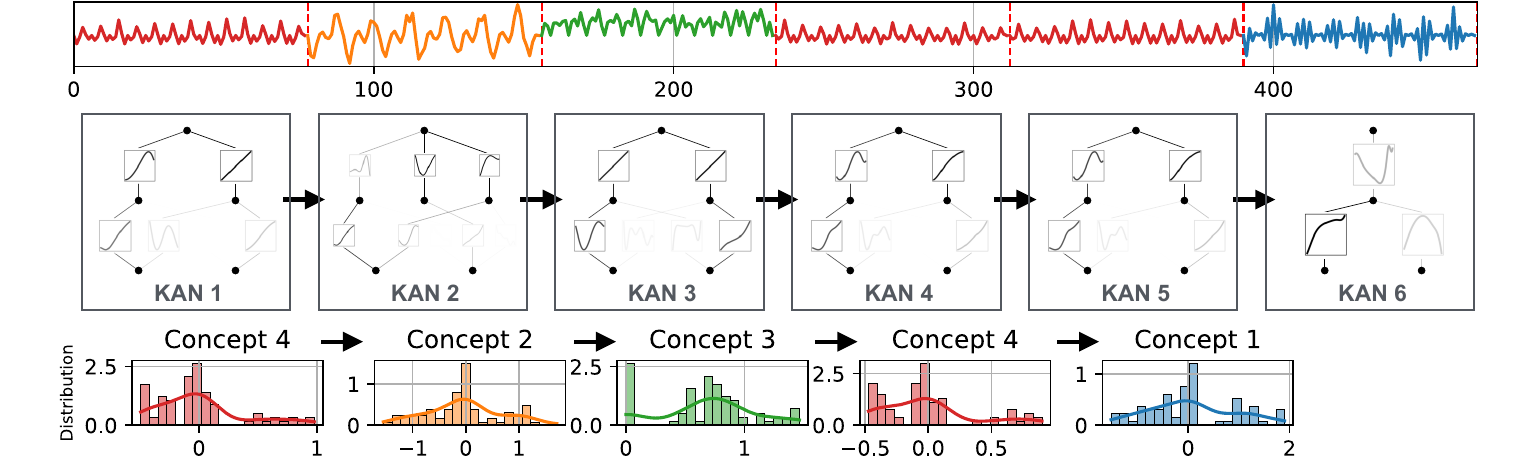}}

\caption{Modeling power of T-KAN for univariate time series: training Temporal KAN (T-KAN) and detecting concept drift}
\label{Fig2}
\vspace{-15pt}
\end{figure}
\subsection{T-KAN}
Time Series Kolmogorov-Arnold Networks (T-KAN) are designed to handle univariate time series data. The primary objective of T-KAN is to simultaneously predict future values and detect as well as track concept drift. 

\subsubsection{Architecture}

The architecture of T-KAN is based on the original KAN framework, with modifications to handle temporal dependencies in the data. T-KAN employs a two-layer (2-depth) network structure where each layer consists of spline-parametrized univariate functions. These functions model the relationships between consecutive time steps, allowing the network to adaptively learn the temporal patterns within the data. 

Formally, the output of T-KAN at time step \( t \), denoted as \( \hat{S}_{t+T} \), is given by:

\begin{equation}
\hat{S}_{t+T} = \sum_{q=1}^{2n+1} \Phi_q \left( \sum_{p=1}^{h} \varphi_{q,p}(S_{t-h+p}) \right),
\end{equation}

where \( S_{t-h+p} \) represents the past observations at previous time steps, and \( \varphi_{q,p} \) are the spline-parametrized univariate functions. Each \( \Phi_q \) and \( \varphi_{q,p} \) is adaptively learned during the training process, allowing T-KAN to effectively model complex, nonlinear temporal dependencies.

As illustrated in Fig. \ref{Fig2}, we use a sliding window to traverse the time series. This approach creates input-output pairs from each window, enabling the training of the KAN model. For simplicity, we use two historical time steps to predict the next time step in our example. Different KAN structures \& activation functions represent different concepts, and by observing the variations in KAN, we can identify concept drift. For instance, the learnable activation functions in KAN1, KAN4, and KAN5 exhibit consistent behavior, indicating that they are under the same concept or pattern. This ensemble of evolving KAN models constitutes our Temporal KAN (T-KAN), which effectively captures and adapts to concept drift in time series data.

\subsubsection{Symbolic Regression for Interpretability}
Symbolic regression is incorporated into T-KAN to enhance interpretability by fitting mathematical expressions to the learnable activation functions. This approach allows us to generate human-readable models that explain the underlying patterns in the data. This feature is particularly useful for understanding how past observations influence future predictions, making T-KAN not only accurate but also transparent, see Fig. \ref{Fig1}.

\subsection{MT-KAN}

Multivariate Time Series Kolmogorov-Arnold Networks (MT-KAN) are designed to handle multivariate time series data. The primary objective of MT-KAN is to uncover and leverage the complex relationships among multiple variables to improve forecasting performance.

\begin{figure}[t]
\centerline{\includegraphics[width=0.7\linewidth]{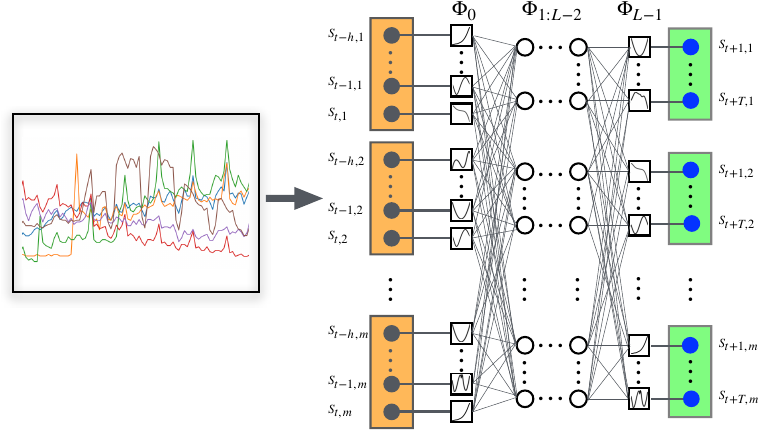}}

\caption{MT-KAN architecture for multivariate time series.}
\label{Fig3}
\vspace{-15pt}
\end{figure}

\subsubsection{Architecture}

The architecture of MT-KAN extends the KAN framework to multivariate settings by modeling interactions between multiple time series. Similar to T-KAN, MT-KAN uses spline-parametrized univariate functions to capture temporal dependencies. However, it also includes mechanisms to model cross-variable interactions, allowing it to exploit the relationships among different variables. As illustrated in Fig. \ref{Fig3}, we simply flatten and stack the historical steps of each variable as inputs. These inputs are then processed through \( L \) layers of KAN, resulting in the forecast horizon for each variable.

Formally, the output of MT-KAN at time step \( t+T \), denoted as \( \hat{\mathbf{S}}_{t+T} \), is given by $\hat{\mathbf{S}}_{t+T} = \sum_{q=1}^{2n+1} \Phi_q \left( \sum_{p=1}^{h} \sum_{k=1}^{m} \varphi_{q,p,k}(\mathbf{S}_{t-h+p,k}) \right),$, where \( \mathbf{S}_{t-h+p,k} \) represents the past observations of the \( k \)-th variable at previous time steps, and \( \varphi_{q,p,k} \) are the spline-parametrized univariate functions. This structure allows MT-KAN to effectively capture both temporal dependencies and cross-variable interactions within the data. 

\subsubsection{Modeling Cross-Variable Interactions}
MT-KAN explicitly models the interactions between different variables in a multivariate time series. This is achieved by incorporating additional parameters and functions that capture the dependencies between variables. By doing so, MT-KAN can leverage the information from multiple time series to make more accurate predictions, thereby enhancing the overall forecasting performance. Specifically, the output of each node not only relies on the historical data of the corresponding time series but also utilizes the historical steps of all variables to make predictions. Each node's output can be traced back to its relationships with all variables and all steps through learnable activation functions.

\section{Experiments}
\subsection{Data}
We utilized a financial time series dataset for our experiments due to its complex and often unpredictable nature, characterized by high volatility and a lack of consistent periodicities. This dataset comprises daily OHCLV (open, high, close, low, volume) data spanning from January 4, 2012, to June 22, 2022\footnote{\url{https://ca.finance.yahoo.com/}}. Our primary objective was to predict the implied volatility of each stock. Since true volatility cannot be directly observed, we approximated it using an estimator based on realized volatility. The conventional volatility estimator is defined as: $\mathcal{V}_t=\sqrt{\sum_{t=1}^n (r_t)^2}$, where $r_t=\ln (c_t/c_{t-1})$ and $c_t$ represents the closing price at time $t$. 
\begin{figure}[t]
\centerline{\includegraphics[width=0.8\linewidth]{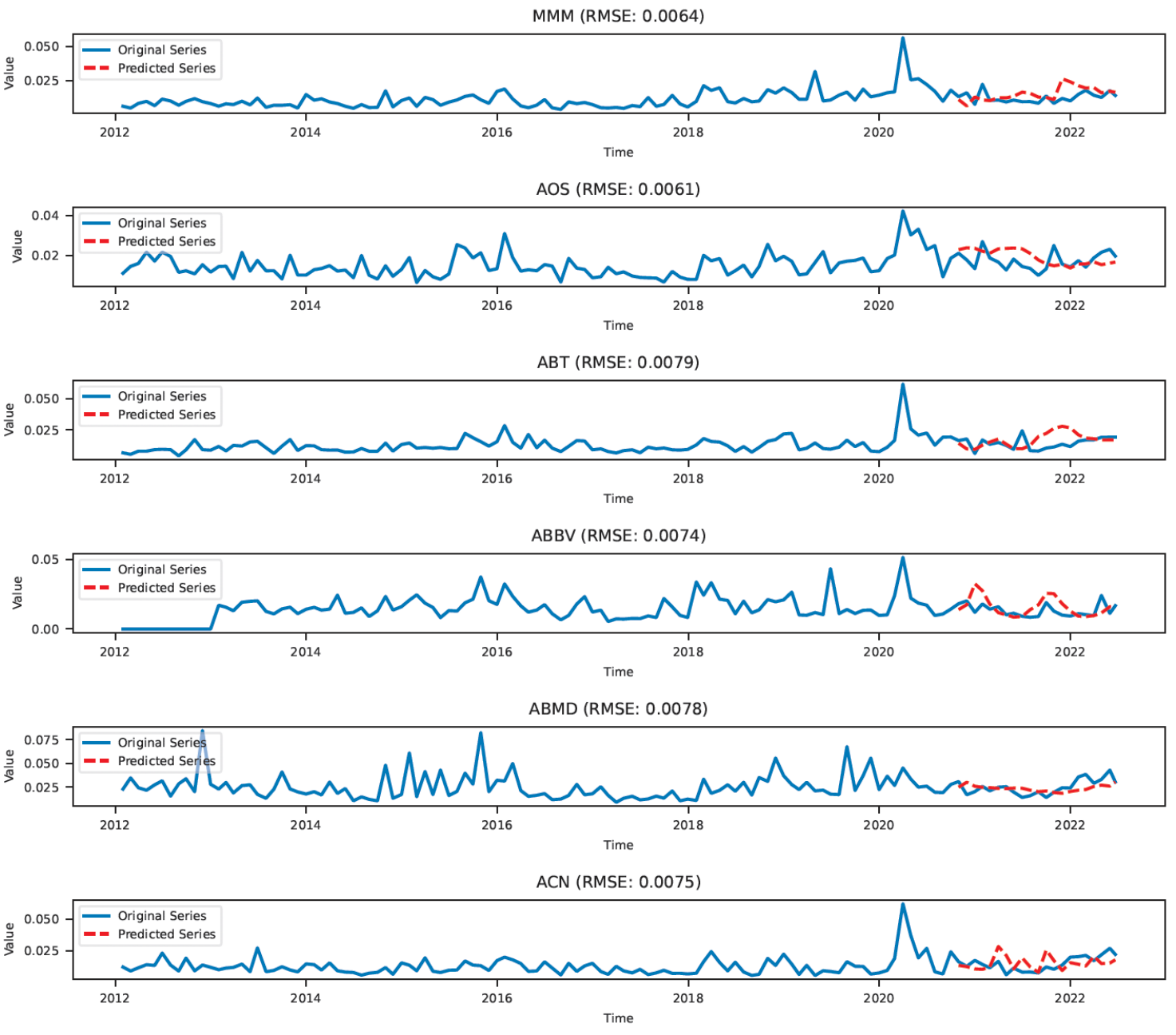}}
\vspace{-10pt}
\caption{Predicted results using a simple T-KAN: true (blue) and forecasted (red) values for the first six series.}
\label{Fig4}
\vspace{-15pt}
\end{figure}
\subsection{Experimental Setup and Baselines}

Both TS-KAN and MT-KAN are implemented using a spline-based parameterization for the univariate functions. For the stock datasets, we use a 21-step window, with four such windows (totaling 84 steps) as inputs to predict the volatility for the next 21 days. We adopt the network structure from the original KAN paper, utilizing a simple two-layer (the input layer is not accounted as a layer per se) architecture with only 5 hidden neurons, i.e., [84, 5, 21]. For MT-KAN, we group 5 variables together for input, resulting in a structure of [84*5, 5, 21*5]. This configuration allows MT-KAN to model and predict the interactions between multiple time series more effectively, especially for stock with \( m = 503 \) variables. We set the training steps to 20 iterations, followed by a pruning step with a threshold of \(5 \times 10^{-2}\), and then another 20 training iterations. This ensures that the models can effectively learn and adapt to the data while maintaining computational efficiency.

We compare our models against classical models, including MLP \cite{kruse2022multi}, RNN \cite{salehinejad2017recent}, and LSTM \cite{graves2012long}, rather than the most recent deep learning methods. Given the early stage of KAN development, it is fair to compare it as a potential alternative to MLPs and other classical methods. This strategy allows us to better evaluate the performance improvements brought by KAN relative to well-established architectures in time series forecasting. The evaluation metrics used in this experiment include Mean Squared Error (MSE), Mean Absolute Error (MAE), and Root Mean squared Error (RMSE).

\subsection{Results and Analysis}
In this subsection, we present the results of our experiments. The results are summarized in Table \ref{tab:comparison_results}. The best results are highlighted in bold, and the suboptimal results are underlined. As shown in Table \ref{tab:comparison_results}, both T-KAN and MT-KAN achieve competitive results. Specifically, T-KAN, with only a single layer of 5 hidden neurons, demonstrates efficiency and robustness comparable to other models that use additional hidden layers or increase the number of hidden neurons. This indicates that even with a simpler architecture, T-KAN can achieve results close to those of more complex models. MT-KAN leverages the nonlinear relationships in multivariate time series to improve prediction accuracy compared to T-KAN. 

\begin{table}[h]
\vspace{-10pt}
    \centering
    \caption{Comparison of forecasting performance between T-KAN, MT-KAN, and baseline models.}
    \vspace{-5pt}
    \resizebox{0.48\textwidth}{!}{%
    \begin{tabular}{lccccc}
        \toprule
        Model & Configuration & MSE & MAE & RMSE & Parameters \\
        \midrule
        MLP & [84,5,21] & 0.0465 & 0.1774 & 0.2141 & 551 \\
        MLP & [84,50,21] & 0.0002 & 0.0122 & 0.0157 & 5321 \\
        MLP & [84,200,21] & \underline{8.92e-5} & 0.0072 & 0.0088 & 21221 \\
        MLP & [84,5,5,21] & 0.0504 & 0.1798 & 0.2230 & 581 \\
        MLP & [84,50,50,21] & 0.0001 & 0.0103 & 0.0130 & 7871 \\
        RNN & [84,5,21] & 0.0541 & 0.1737 & 0.2282 & 166 \\
        RNN & [84,50,21] & 0.0001 & 0.0079 & 0.0098 & 3721 \\
        RNN & [84,200,21] & \underline{8.03e-5} & \underline{0.0069} & 0.0083 & 44821 \\
        RNN & [84,5,5,21] & 0.0497 & 0.1691 & 0.2185 & 226 \\
        RNN & [84,50,50,21] & 0.0001 & 0.0079 & 0.0098 & 8821 \\
        LSTM & [84,5,21] & 0.0132 & 0.0737 & 0.1105 & 286 \\
        LSTM & [84,50,21] & \underline{6.69e-5} & \underline{0.0066} & \underline{0.0078} & 11671 \\
        LSTM & [84,200,21] & \underline{6.52e-5} & \underline{0.0064} & \underline{0.0075} & 166621 \\
        LSTM & [84,5,5,21] & 0.0136 & 0.0777 & 0.1124 & 526 \\
        LSTM & [84,50,50,21] & \underline{6.67e-5} & \underline{0.0066} & \underline{0.0076} & 32071 \\
        T-KAN & [84,5,21] & \underline{6.91e-5} & \underline{0.0069} & \underline{0.0078} & 193 \\
        MT-KAN & [84*5,5,21*5] & \textbf{6.37e-5} & \textbf{0.0062} & \textbf{0.0075} & 2132 \\
        \bottomrule
    \end{tabular}}

    \label{tab:comparison_results}
    
\end{table}
Furthermore, the efficiency of KAN-based models in terms of parameter count is evident. For instance, T-KAN has only 193 parameters, significantly fewer than the comparable LSTM model with [84, 50, 21] configuration, which has 11,671 parameters. Similarly, MT-KAN, with its [84*5, 5, 21*5] configuration, manages to outperform other models while using 2,132 parameters. This indicates that KAN-based models can achieve high accuracy with fewer parameters.

We visualized the original series (in blue) and the predicted results (in red) for the first 6 stocks (Nasdaq: MMM, AOS, ABT, ABBV, ABMD, ACN) from the dataset. As shown in Fig.\ref{Fig4}, despite having only 2 layers (including the output layer), 5 hidden neurons, and training for only 40 steps, the predicted values align closely with the original time series\footnote{More results can be found in \url{https://anonymous.4open.science/api/repo/CIKM-7A4B/file/predictions_with_simpleTKAN.pdf?v=ff1d927e}}.


\section{Conclusion}
In this paper, we explored the application of Kolmogorov-Arnold Networks (KAN) to time series forecasting and introduced two specialized variants: T-KAN and MT-KAN. We demonstrated that KAN, offers a high degree of interpretability and predictive accuracy on time series. By replacing traditional linear weights with spline-parametrized univariate functions, KAN dynamically adapts to data patterns, providing a robust framework for forecasting tasks.

\noindent \textbf{Limitation.} Despite the advantages, the implementation of KAN is typically 10 times slower to train compared to MLPs with the same number of parameters. This inefficiency arises because KAN' diverse activation functions cannot fully leverage batch processing. A potential solution is to group activation functions into "heads" that share the same function, balancing between KAN and MLPs. For fast training needs, MLPs are preferable, but KAN is valuable for tasks requiring interpretability and complex relationship modeling.

\noindent \textbf{Future work.} Future work will explore integrating KAN with architectures like RNN, LSTM, and Transformer to enhance interpretability and flexibility. Leveraging adaptive sequence segmentation could improve KAN’s handling of non-stationary data by dynamically partitioning time series based on detected patterns. Additionally, optimizing training speed through parallel processing, efficient batch handling, and streamlined spline calculations will be crucial for making KAN practical for real-time applications.

\bibliographystyle{unsrt}  
\bibliography{references}

\end{document}